\documentclass[lettersize,journal]{IEEEtran}
\usepackage{xcolor}
\usepackage{amsmath}
\usepackage{amsfonts}
\usepackage[noend]{algpseudocode}
\usepackage[ruled,linesnumbered]{algorithm2e}
\SetArgSty{textup}
\usepackage{hyperref}
\usepackage{amssymb}
\usepackage{mathtools}
\usepackage{mathrsfs}
\usepackage{multirow}
\usepackage{indentfirst}
\usepackage{booktabs}
\usepackage{tabularx}
\usepackage{threeparttable}
\usepackage{subcaption}
\usepackage{array}
\usepackage{textcomp}
\usepackage{stfloats}
\usepackage{url}
\usepackage{verbatim}
\usepackage{graphicx}
\usepackage{amsthm}
\usepackage{cite}
\usepackage{diagbox}
\usepackage{makecell}
\usepackage{tikz}

\def\BibTeX{{\rm B\kern-.05em{\sc i\kern-.025em b}\kern-.08em
    T\kern-.1667em\lower.7ex\hbox{E}\kern-.125emX}}
\usepackage{balance}

\usepackage{amsthm}
\theoremstyle{definition}

\newtheorem{proposition}{\bf Proposition}

\usepackage{kotex}

\usepackage{xcolor}

\begin{document}

\title{BPMP-Tracker: A Versatile Aerial Target Tracker Using Bernstein Polynomial Motion Primitives}

\author{Yunwoo Lee$^{1}$, Jungwon Park$^{2}$, Boseong Jeon$^{3}$,  Seungwoo Jung$^{1}$, and H. Jin Kim$^{1}$
\thanks{$^{1}$The authors are with the Department of Mechanical and Aerospace Engineering, Seoul National University, Seoul, South Korea (e-mail: snunoo12@snu.ac.kr; tmddn833@snu.ac.kr; hjinkim@snu.ac.kr).}\
\thanks{$^{2}$The author is with Department of Mechanical System Design Engineering, Seoul National University of Science and Technology (SEOULTECH), Seoul, South Korea (e-mail: jungwonpark@seoultech.ac.kr)}\
\thanks{$^{3}$The author is with Samsung Research, Samsung Electronics, Seoul, South Korea (e-mail: junbs95@gmail.com).}
}

\markboth{}%
{}

\maketitle
\begin{abstract}
This letter presents a versatile trajectory planning pipeline for aerial tracking. The proposed tracker is capable of handling various chasing settings such as complex unstructured environments, crowded dynamic obstacles and multiple-target following. Among the entire pipeline, we focus on developing a predictor for future target motion and a chasing trajectory planner. For rapid computation, we employ the \textit{sample-check-select} strategy: modules sample a set of candidate movements, check multiple constraints, and then select the best trajectory. Also, we leverage the properties of Bernstein polynomials for quick calculations. The prediction module predicts the trajectories of the targets, which do not overlap with static and dynamic obstacles. Then the trajectory planner outputs a trajectory, ensuring various conditions such as occlusion and collision avoidance, the visibility of all targets within a camera image and dynamical limits. We fully test the proposed tracker in simulations and hardware experiments under challenging scenarios, including dual-target following, environments with dozens of dynamic obstacles and complex indoor and outdoor spaces.
\end{abstract}


\IEEEpeerreviewmaketitle

\section{Introduction}
\label{sec:introduction}
\IEEEPARstart{A}{dvancements} in motion planning for micro drones \cite{minco} promote developments of autonomous aerial tracking and its utilization in various applications, including videography \cite{cinempc} and surveillance \cite{surveillance}. Target following tasks require additional capabilities to predict the future motion of the target amidst complex obstacles and calculate the trajectory to ensure that obstacles do not occlude the target, as well as to reflect the factors commonly considered in general trajectory planning, such as drone safety, path quality and actuator limits. 

The visibility of the targets has been considered in recent trajectory planning research for specific mission settings such as single- \cite{fast_tracker,boseong_icra,bonatti} or dual- \cite{dual_target} target tracking and static unstructured \cite{visibility_aware,elastic_tracker,bonatti2} or dynamic structured (\textit{e.g.,} ellipsoid) \cite{nageli,multi-convex} environments. However, the number of targets of interest may vary depending on the situation, and there may be moving objects other than the targets. Such limitations necessitate the development of a chasing planner that can be applied to various settings. 

Furthermore, to respond to unforeseen atypical obstacles and changes in the motion of multiple dynamic objects, prediction and planning should be performed fast enough. Given the limited computation resources on the drone, the planner should efficiently handle tasks of any difficulty.


This paper presents an online target-following planner that can handle various scenarios. We focus on target trajectory prediction and drone trajectory planning, utilizing the \textit{sample-check-select} strategy to effectively handle various non-convex constraints. A large set of polynomial motion primitives are rapidly \textit{sampled}, and using the properties of Bernstein polynomials, reliable feasibility \textit{checks} for dynamic limits, collision and occlusion avoidance and camera field-of-view (FOV) constraints are quickly performed. Subsequently, the primitive with the best score is \textit{selected} as the final trajectory. This approach enables multi-thread parallel computation, and from \textit{sample} to the \textit{select}, it consumes only a few microseconds to handle a single primitive, resulting in low computational burdens to the entire pipeline. Extensive tests with various challenging scenarios are performed to show the efficiency and effectiveness of the tracking planner.
\begin{figure}[t!]
\centering
\includegraphics[width=1.0\linewidth]{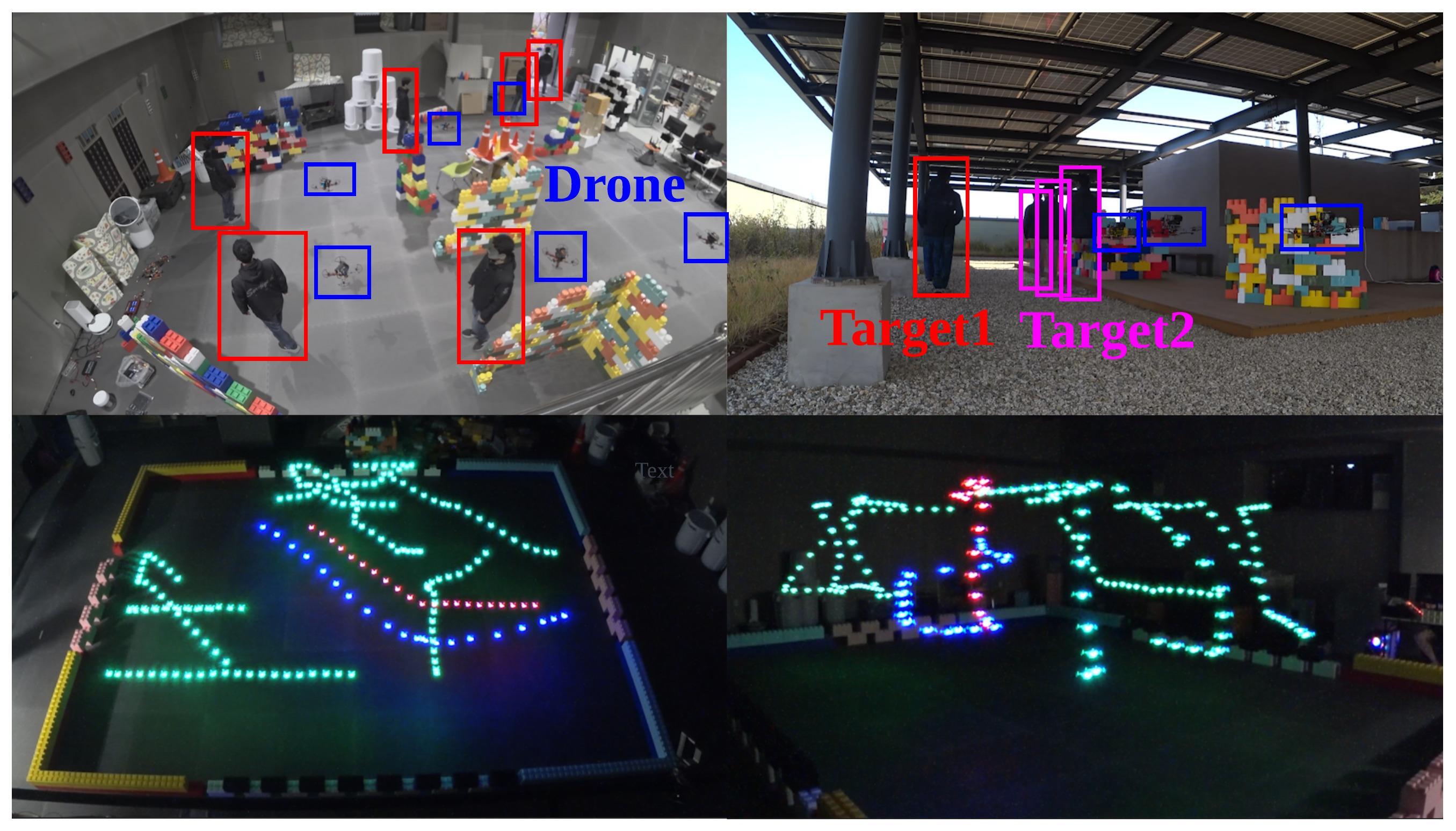}
\caption{Various target chasing scenarios. A chaser (blue) follows the targets (red, magenta) in unstructured spaces (top) and among dynamic obstacles (green, bottom).}
\label{fig:thumbnail}
\end{figure}

The main contributions of our work are as follows.
\begin{itemize}
    \item Versatile target tracking planner that can be adopted for both unstructured- and dynamic- obstacle environments, capable of handling various numbers of targets.
    \item Efficient target prediction and trajectory planning algorithm scalable to dense and complex environments.
    \item Extensive validations of our approaches, including challenging and large-scale simulations and actual drone experiments, inclusive of scenarios with dozens of dynamic obstacle environments and a scenario starting from indoor and ending in outdoor space.
\end{itemize}
\section{Related Works}
\label{sec:related_works}
We briefly categorize planning research for target tracking.

\textbf{Empty environments}: Research on motion planning to track a target in a free space has been extensively reported. \cite{control1,control2,control3} propose tracking controllers to prevent the target from moving out of a camera image. 
Their works have succeeded in preserving the visibility of the target, but they are only adoptable in obstacle-free spaces.

\textbf{Unstructured-but-static} \textbf{environments}: Various research has proposed methods for target tracking in obstacle environments. Regarding drone trajectory generation, \cite{boseong_icra} proposes a visibility metric to acquire optimal viewpoints in unstructured space. It effectively measures the degree of target visibility against general obstacles but it is possible to meet occlusion between viewpoints. \cite{elastic_tracker,visibility_aware} formulate target occlusion avoidance as hard constraints and optimize trajectories using techniques derived from \cite{minco}, which converts constrained optimization into unconstrained one. They succeed in collision and occlusion avoidance during flight, but they do not consider dynamic obstacles.

\textbf{Dynamic-but-structured environments}: There are approaches addressing aerial tracking in dynamic environments using soft constraints \cite{nageli} and hard constraints \cite{multi-convex}.
However, unstructured obstacles are not handled in the existing works that address target visibility among moving ellipsoidal obstacles.
Moreover, a collision between targets and obstacles is not reflected.
When the paths of the targets and obstacles overlap, their optimization inevitably leads to failure.

\textbf{Target prediction}: Regarding target motion prediction, \cite{fast_tracker} and \cite{fast_tracker2} suggested methods predicting the motion of moving objects using polynomial regression. However, since there is no consideration of obstacles, it may produce prediction results that pass through occupied spaces. \cite{boseong_icra,dual_target} and \cite{qp_chaser} reflect obstacles in the prediction method, but their methods are applicable only to either unstructured-but-static or dynamic-but-structured obstacles. 

\textbf{Multiple-target following}: Some works deal with multiple-target following using a single drone \cite{dual_target,qp_chaser,multi_target}. They aim to maximize information of the targets in motion making; however, their performance is limited to specific conditioned environments, like the other studies above.

\section{Problem Formulation}
\label{sec:problem_formulation}
\begin{figure}[t!]
\centering
\includegraphics[width=0.94\linewidth]{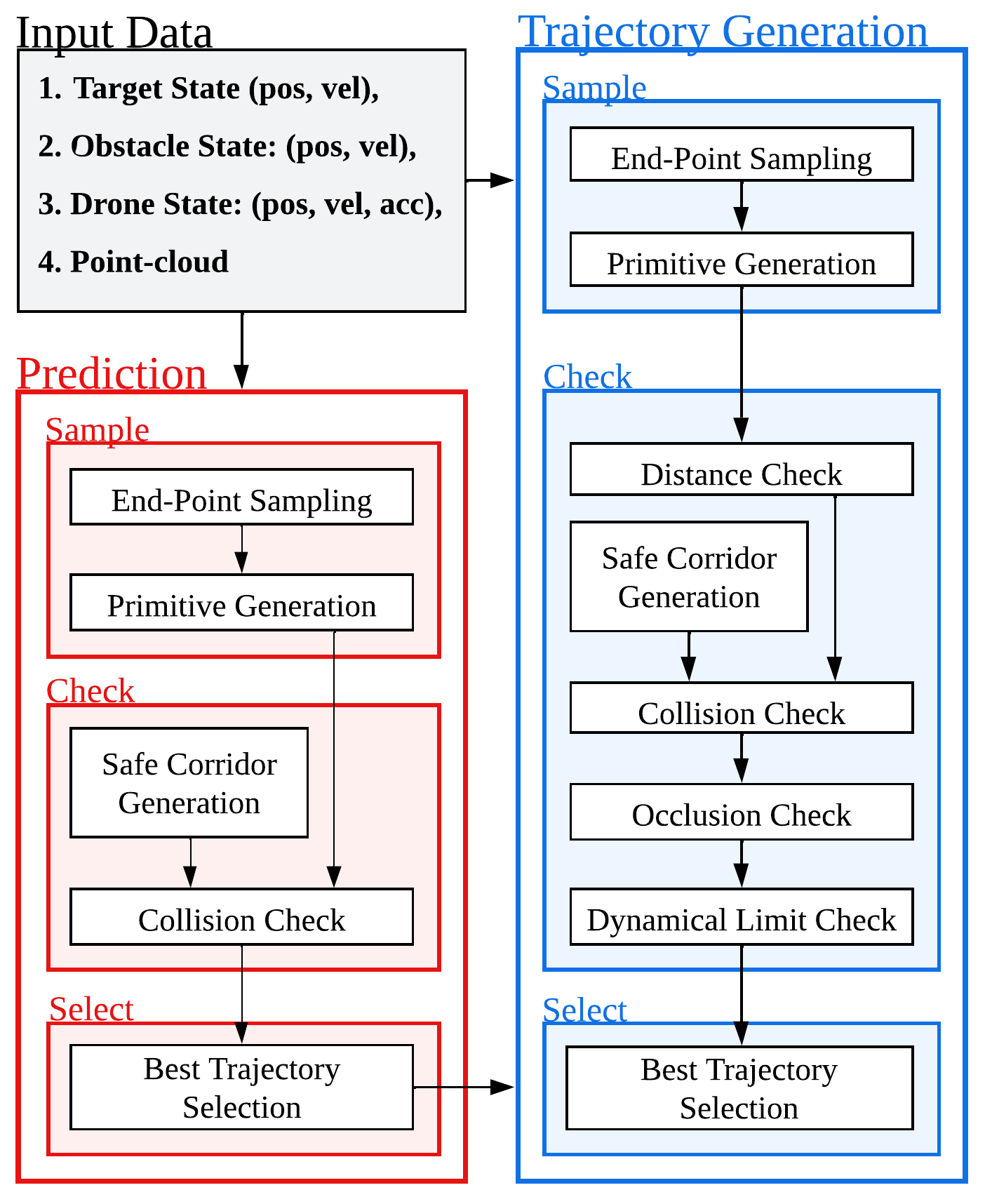}
\caption{Overview of the chasing pipeline.}
\label{fig:pipeline}
\end{figure}

In this section, we formulate a trajectory planning problem for a following drone, which keeps target visibility. We suppose that $N_{q}$ targets are in a 3-dimensional space $\mathcal{X}\subset\mathbb{R}^{3}$ with static and dynamic obstacles. Our goal is to predict the non-collision future movement of targets and generate a continuous-time trajectory so that the drone 1) maintains desired shooting distance from the targets, 2) avoids collisions, 3) keeps visibility, and 4) satisfies dynamical feasibility. The whole pipeline is summarized in Fig. 
\ref{fig:pipeline}.
\subsection{Environments}
\label{subsec:environments}
The environment $\mathcal{X}$ consists of obstacle space $\mathcal{O}(t)$, target space $\mathcal{T}(t)$, and the drone. $\mathcal{O}(t)$ is divided into spaces occupied by static and dynamic obstacles, $\mathcal{O}_{s}$ and $\mathcal{O}_d(t)$, respectively, and $\mathcal{T}(t)$ is a space occupied by targets. Based on the spaces, we define three free spaces for chasing missions.
\begin{equation}
    \label{eq:free_spaces}
    \mathcal{F}_{s} = \mathcal{X}\setminus \mathcal{O}_{s}, \ 
     \mathcal{F}_{d}(t) = \mathcal{F}_{s}\setminus \mathcal{O}_{d}(t),\
     \mathcal{F}_{q}(t) = \mathcal{F}_{d}(t)\setminus \mathcal{T}(t)
\end{equation}
\subsection{Assumptions}
\label{subsec:assumptions}
We assume that moving objects, including target and dynamic obstacles, are ellipsoidal and maintain their shape when they move: 
$\mathcal{E}(\textbf{q}(t),Q) = \{\textbf{x}(t)\in \mathbb{R}^{3}\ | \ \|\textbf{x}(t)-\textbf{q}(t)\|_{Q}\leq1,\ Q\in \mathbb{S}^{++}$\}.
For mathematical simplicity, we assume all ellipsoids to be axis-aligned, and extending to a rotated ellipsoid is straightforward. Also, we assume that the targets do not move in a jerky way.
\subsection{Trajectory Representation}
\label{subsec:trajectory_representation}
Thanks to the differential flatness of quad-rotor dynamics, we represent trajectories with polynomial functions of time $t$. This paper employs Bernstein basis \cite{bebot} to express the polynomials, and we leverage the following properties of Bernstein polynomials.\\
\textit{P1) Nonenegativeness:} The Bernstein basis is non-negative throughout its time interval. All nonnegative Bernstein coefficients result in the polynomial being nonnegative.\\
\textit{P2) Multiplication:} Multiplication of two Bernstein polynomials yields a Bernstein polynomial.\\
\textit{P3) Convex hull:} Bernstein polynomial lies within the convex hull defined by its Bernstein coefficients.\\
Using the above properties, feasibility checks can be performed efficiently using Bernstein coefficients with few arithmetic operations without conducting feasibility checks for finely discretized time steps.
The trajectory of the drone, $\textbf{x}_{c}(t) \in \mathbb{R}^{3}$, is represented as follows:
\begin{equation}
    \textbf{x}_{c}(t) = \textbf{C}^{\top}\textbf{b}_{n_c}(t),\ t\in [0,T]
    \label{eq:representation}
\end{equation}
where $n_c$ is the degree of the polynomial, $\textbf{b}_{n_c}(t)\in \mathbb{R}^{(n_{c}+1)\times 1}$ is a vector that consists of $n_c$-th order Bernstein bases for time horizon $[0,T]$, and $\textbf{C}\in \mathbb{R}^{(n_c+1) \times 3}$ is a coefficient matrix.

Likewise, the trajectories of the $i$-th targets, $\textbf{x}_{q}^{i}(t)$, the $j$-th dynamical obstacles, $\textbf{x}_{o}^{j}(t)$ are represented in the same manner. In this paper, we omit the superscripts $i$ and $j$ to express an arbitrary single object.
\subsection{Mission Description}
\label{subsec:mission_description}
\subsubsection{Target Distance}
\label{subsubsec:target_distance}
To prevent getting too close or far from the target, the distance constraint is formulated as follows:
\begin{equation}
    \label{eq:distance_target}
    d_{\min} \leq \|\textbf{x}_{c}(t)-\textbf{x}_{q}(t)\|_{2}\leq d_{\max}, \ 
    \forall t \in [0,T]
\end{equation}
where $d_{\min}$ and $d_{\max}$ are the minimum and maximum values of the desired distance between the targets and the drone, respectively.
\subsubsection{Collision Avoidance}
\label{subsubsec:collision_avoidance}
In order not to collide with the environment, we formulate the following constraints.
\begin{subequations}
\label{eq:collision_avoidance}
\begin{align}
    \label{subeq:target_collision}
    & \mathcal{E}(\textbf{x}_{q}(t),diag(r_{qx}^{-2},r_{qy}^{-2},r_{qz}^{-2}))\subset \mathcal{F}_{d}(t), \forall{t}\in [0,T] \\
    \label{subeq:drone_collision}
    &\mathcal{E}(\textbf{x}_{c}(t),r_{c}^{-2}I_{3})\subset \mathcal{F}_{q}(t), \forall{t}\in [0,T]
\end{align}
\end{subequations}
where $[r_{qx},r_{qy},r_{qz}]$ are lengths of the semi-axes of the smallest ellipsoid enclosing a target, $diag(\cdot,\cdot,\cdot)$ is a diagonal matrix, $r_{c}$ is the radius of the drone, and $I_{3}$ is the $3 \times 3$ identity matrix. (\ref{subeq:target_collision}) represents that the target does not collide with obstacles $\mathcal{O}(t)$, and (\ref{subeq:drone_collision}) prevents collision of the drone against the $\mathcal{O}(t)\cup \mathcal{T}(t)$.


\subsubsection{Target Visibility}
\label{subsubsec:target_visibility}
To avoid target occlusion by obstacles, the \textit{Line-of-Sight} between the drone and the target, $\mathcal{L}(\textbf{x}_{c}(t),\textbf{x}_{q}(t)) = \{\textbf{x}(t)\ | \ \textbf{x}(t)= (1-\epsilon)\textbf{x}_{c}(t)+ \epsilon \textbf{x}_{q}(t), \forall \epsilon \in [0,1] \}$, should not intersect with the obstacle space $\mathcal{O}(t)$.
\begin{equation}
\label{eq:occlusion_avoidance}
    \mathcal{L}(\textbf{x}_{c}(t),\textbf{x}_{q}(t))\subset \mathcal{F}_{d}(t), \forall{t}\in [0,T]
\end{equation}
Additionally, while following multiple targets, the drone should simultaneously see the targets with a limited field of view of a camera, $\theta_{f}\in[0,\pi]$. The angle made by the two \textit{Line-of-Sights}, $\mathcal{L}(\textbf{x}_{c}(t),\textbf{x}_{q}^{i}(t))$ and $\mathcal{L}(\textbf{x}_{c}(t),\textbf{x}_{q}^{j}(t)), i\neq j$, should not exceed $\theta_{f}$.
\begin{equation}
    \label{eq:fov_constraints}
    \frac{(\textbf{x}_{q}^{i}(t)-\textbf{x}_{c}(t))^{\top}(\textbf{x}_{q}^{j}(t)-\textbf{x}_{c}(t))}{\|\textbf{x}_{q}^{i}(t)-\textbf{x}_{c}(t)\|_{2}\|\textbf{x}_{q}^{j}(t)-\textbf{x}_{c}(t) \|_{2}}\geq \cos{\theta_{f}}, \forall t\in[0,T]
\end{equation}

\subsubsection{Dynamical Limits}
\label{subsubsec:dynamical_limits}
Due to the actuator limits of the drone, the generated trajectory should not exceed the maximum velocity, $v_{\max}$, and acceleration, $a_{\max}$:
\begin{subequations}
\label{eq:dynamical_limits}
\begin{align}
& \|\dot{\textbf{x}}_{c}(t)\|_{2}\leq v_{\max},\ \forall t \in [0,T]\\
& \|\ddot{\textbf{x}}_{c}(t)\|_{2}\leq a_{\max},\ \forall t \in [0,T]
\end{align}
\end{subequations}
Also, a high rate of yaw change causes motion blur and hinders the performance of onboard algorithms (e.g., localization, target detection); therefore, we constrain the yaw rate:
\begin{equation}
    \label{eq:yaw_rate}
    |\dot{\psi}(t)|\leq \dot{\psi}_{\max}\ \forall t \in [0,T].
\end{equation}

\section{Target Trajectory Prediction}
\label{sec:target_trajectory_prediction}
This section introduces the calculation of future target trajectories. First, we sample a bundle of motion primitives and filter out the primitives that collide with obstacles. Then, the best trajectory is selected. 
\subsection{Target Primitive Sampling}
\label{subsec:target_primitive_sampling}
Based on the current position $\textbf{x}_{q0}$ and velocity $\dot{\textbf{x}}_{q0}$, the predictor samples dynamically plausible primitives.\\
\textit{Nearly Constant Velocity Model (NCVM)}: We modify the constant velocity model to calculate possible target trajectories. With the assumption that targets do not move in a jerk way, the primitive is generated to smoothly interpolate the initial position $\textbf{x}_{q0}$ and sampled reachable positions at time $T$, $\textbf{x}_{q}^{f}$, considering the current velocity $\dot{\textbf{x}}_{q0}$. 
\begin{equation}
    \label{eq:nearly_constant_velocity_model}
    \begin{aligned}
    &\underset{\textbf{a}_{q}(t)}{\text{min}} && \frac{1}{T}\int_{0}^{T}\|\textbf{a}_{q}(t)\|^{2}_{2}dt \\
    &\text{s.t.} &&  \frac{d}{dt}
    \begin{bmatrix}
    \textbf{x}_{q}(t)\\ \dot{\textbf{x}}_{q}(t)
    \end{bmatrix}
     = F_{q}
    \begin{bmatrix}
    \textbf{x}_{q}(t)\\ \dot{\textbf{x}}_{q}(t)
    \end{bmatrix} + G_{q}\textbf{a}_{q}(t),\\
     & \ &&F_q =
\begin{bmatrix}
    0_{3} & I_{3}\\
    0_{3} & 0_{3}
\end{bmatrix}
, \ G_q = 
\begin{bmatrix}
    0_{3} \\
    I_{3}
\end{bmatrix},\\
& \ && \textbf{x}_{q}(0)= \textbf{x}_{q0}, \ \dot{\textbf{x}}_{q}(0) = \dot{\textbf{x}}_{q0}, \ \textbf{x}_{q}(T)=\textbf{x}_{q}^{f},\\
& \ && \textbf{x}_{q}^{f} \sim \mathcal{N}(\textbf{x}_{q0}+\dot{\textbf{x}}_{q0}T,P_{q}(T))
    \end{aligned}
\end{equation}
where $0_{3}$ is a $3\times3$ zero matrix, $P_{q}(T)\in \mathbb{R}^{3\times3}$ is the error covariance matrix at time $T$, which is calculated from state estimator (\textit{e.g.,} extended Kalman filter), and $\mathcal{N}(\textbf{a},B)$ represents a gaussian distribution with mean $\textbf{a}$ and variance $B$. After sampling $\textbf{x}_{q}^{f}$, the optimal control problem in (\ref{eq:nearly_constant_velocity_model}) is solved in closed form, and the optimal trajectory $\textbf{x}_{q}(t)$ can be written in cubic Bernstein polynomial form. 
The coefficients of the solution are as follows:
\begin{equation}
\textbf{Q}^{\top} = [\textbf{x}_{q0}, \ \textbf{x}_{q0}+\frac{T}{3}\dot{\textbf{x}}_{q0},\ \frac{1}{2}\textbf{x}_{q0}+\frac{1}{2}\textbf{x}_{q}^{f}+\frac{T}{6}\dot{\textbf{x}}_{q0},\ \textbf{x}_{q}^{f}]
    \label{eq:primitive_target}
\end{equation}

\subsection{Collision Check}
\label{subsec:collision_check_target}
To select primitives of the target that avoid obstacles, we examine the following conditions.
\begin{subequations}
    \label{eq:collision_check_target}
\begin{align}
    \label{subeq:collision_between_static_target}
    & \mathcal{V}(\textbf{Q})\oplus \mathcal{E}(0,diag(r_{qx}^{-2},r_{qy}^{-2},r_{qz}^{-2}))\subset \mathcal{S}^{q},\\
    \label{subeq:collision_between_dynamic_target}
    & \|\textbf{x}_{q}(t)-\textbf{x}_{o}(t)\|_{Q_{q}^{o}}^{2}- 1\geq0
\end{align}
\end{subequations}
where $\mathcal{V}(\cdot)$ is a set of points in Bernstein coefficients, $\oplus$ denotes a Minkowski-sum,  $\mathcal{S}^{q}$ is safe corridor inflated from seeds $\textbf{x}_{q0}$ in $\mathcal{F}_{s}$ \cite{sfc}, $Q_{q}^{o}=diag((r_{qx}+r_{ox})^{-2},(r_{qy}+r_{oy})^{-2},(r_{qz}+r_{oz})^{-2})$, and $[r_{ox},r_{oy},r_{oz}]$ are lengths of semi-axes of a dynamic obstacle.  
Because $\mathcal{S}^{q}\cap \mathcal{O}_{s}= \emptyset$, the targets moving along $\textbf{x}_{q}(t)$ that satisfy (\ref{subeq:collision_between_static_target}) do not collide with static obstacles. (\ref{subeq:collision_between_dynamic_target}) is a condition to avoid collisions with dynamic obstacles, so the primitives that pass (\ref{eq:collision_check_target}) satisfy (\ref{subeq:target_collision}). 
We utilize the property \textit{P3)} to check (\ref{subeq:collision_between_static_target}) and exploit \textit{P1}) and \textit{P2}) to check (\ref{subeq:collision_between_dynamic_target}). Feasibility checks in drone trajectory planning adopt the same form of constraints as those in (\ref{eq:collision_check_target}). The checks are similarly conducted using \textit{P1)}-\textit{P3)}.

\subsection{Best Prediction Selection}
\label{subsec:best_trajectory_target}
Among the primitives that pass (\ref{eq:collision_check_target}), we select the best future target trajectory. We determine the best trajectory $\textbf{x}_{q}(t)$ as a center primitive among noncolliding primitives. 
\begin{equation}
    \label{eq:best_primitive_selection}
    \min_{i} \sum_{j\neq i}^{|\mathcal{P}_{s}^{q}|} \int_{0}^{T}\| {}^{i}_{s}\textbf{x}_{q}(t)- {}^{j}_{s}\textbf{x}_{q}(t)\|_{2}^{2}dt
\end{equation}
where $\mathcal{P}_{s}^{q}$ is a set of noncolliding primitives, and ${}_{s}^{i}\textbf{x}_{q}(t)$ represents the $i$-th primitives in $\mathcal{P}_{s}^{q}$. Since the integrand in (\ref{eq:best_primitive_selection}) is a Bernstein polynomial, the integral can be calculated by summing the coefficients, which leads to fast computation. A prediction example is shown in Fig. \ref{subfig:target_prediction}.
\section{Chasing Trajectory Planning}
\label{sec:chasing_trajectory_planning}
Given the future trajectory of targets and dynamic obstacles, $\textbf{x}_{q}(t)$ and $\textbf{x}_{o}(t)$, $t\in [0,T]$, and point-cloud $\mathcal{P}$, we calculate the trajectory for the chasing drone. Similar to the target prediction, after sampling motion primitives, various constraint checks are performed, and then the best trajectory is selected.
\subsection{Drone Primitive Sampling}
\label{subsec:drone_primitive_sampling}
Based on the chasing drone's current position $\textbf{x}_{c0}$, velocity $\dot{\textbf{x}}_{c0}$, and acceleration $\ddot{\textbf{x}}_{c0}$, we sample motion primitives for the drone. First, we sample shooting position at time $T$. In a spherical coordinate system centered around the target's terminal points $\textbf{x}_{q}(T)$, samples are taken using the following method:
\begin{equation}
    \label{eq:terminal_point_sampling}
    \begin{aligned}
    &\textbf{x}_{c}^{f} = \textbf{x}_{q}(T)+[r_{c}^{s}\cos{\psi_{c}^{s}}\cos{\phi_{c}^{s}},r_{c}^{s}\cos{\psi_{c}^{s}}\sin{\phi_{c}^{s}},r_{c}^{s}\sin{\psi_{c}^{s}} ]^{\top},\\
    &r_{c}^{s} \sim U[\underbar{$r$}_{c},\bar{r}_{c}],\
    \psi_{c}^{s} \sim U [\underbar{$\psi$}_{c},\bar{\psi}_{c}], \
    \phi_{c}^{s} \sim U [\underbar{$\phi$}_{c},\bar{\phi}_{c}]
    \end{aligned}
\end{equation}
$U$ represents the uniform distribution, and $(\underbar{$r$}_{c},\bar{r}_{c})$, $(\underbar{$\psi$}_{c},\bar{\psi}_{c})$, and $(\underbar{$\phi$}_{c},\bar{\phi}_{c})$ are pairs of the lower and upper bound of distribution of radius, elevation, and azimuth, respectively.

Then we generate primitives based on the minimum jerk problem as follows:

\begin{equation}
    \label{eq:min_jerk_problem}
    \begin{aligned}
    &\underset{\textbf{j}(t)}{\text{min}} && \frac{1}{T}\int_{0}^{T}\|\textbf{j}(t)\|^{2}_{2}dt \\
    &\text{s.t.} && \frac{d}{dt} \begin{bmatrix}
      \textbf{x}_{c}(t) \\ \dot{\textbf{x}}_{c}(t) \\ \ddot{\textbf{x}}_{c}(t)  
    \end{bmatrix} = F_{c}\begin{bmatrix}
      \textbf{x}_{c}(t) \\ \dot{\textbf{x}}_{c}(t) \\ \ddot{\textbf{x}}_{c}(t)  
    \end{bmatrix}+G_{c}\textbf{j}(t) ,\\
    &\ && F_{c} = \begin{bmatrix}
        0_{3} &I_{3} &0_{3} \\
        0_{3} &0_{3} &I_{3} \\
        0_{3} &0_{3} &0_{3}
    \end{bmatrix}, G_{c}=\begin{bmatrix}
        0_{3} \\ 0_{3}\\ I_{3}
    \end{bmatrix},\\
    & \ && \textbf{x}_{c}(0)=\textbf{x}_{c0}, \dot{\textbf{x}}_{c}(0)=\dot{\textbf{x}}_{c0}, \ddot{\textbf{x}}_{c}(0)= \ddot{\textbf{x}}_{c0},\\
    & \ && \textbf{x}_{c}(T)= \textbf{x}_{c}^{f}
    \end{aligned}
\end{equation}
(\ref{eq:min_jerk_problem}) is solved in closed form, and the optimal trajectory $\textbf{x}_{c}(t)$ can be written as a quintic Bernstein polynomial. The coefficients of the solution are as follows:
\begin{equation}
\label{eq:primitive_drone}
\begin{aligned}
\textbf{C}^{\top} = &[\textbf{x}_{c0},\ \textbf{x}_{c0}+\frac{T}{5}\dot{\textbf{x}}_{c0},\ \textbf{x}_{c0}+\frac{2T}{5}\dot{\textbf{x}}_{c0}+\frac{T^{2}}{20}\ddot{\textbf{x}}_{c0},\\
\ & \ \frac{5}{6}\textbf{x}_{c0}+\frac{1}{6}\textbf{x}_{c}^{f}+\frac{13T}{30}\dot{\textbf{x}}_{c0}+\frac{T^{2}}{15}\ddot{\textbf{x}}_{c0},\\
\ & \ \frac{1}{2}\textbf{x}_{c0}+\frac{1}{2}\textbf{x}_{c}^{f}+\frac{3T}{10}\dot{\textbf{x}}_{c0}+\frac{T^{2}}{20}\ddot{\textbf{x}}_{c0},\ \textbf{x}_{c}^{f}]    
\end{aligned}
\end{equation}
\subsection{Feasibility Check}
\label{subsec:feasibility_check_chaser}
\subsubsection{Distance Check}
\label{subsubsec:distance_check}
To keep a suitable distance from the targets, the following conditions are established.
\begin{equation}
    d_{\min}^{2} \leq \|\textbf{x}_{c}(t)-\textbf{x}_{q}(t)\|_{2}^{2}\leq d_{\max}^{2}
    \label{eq:distance_target_divided}
\end{equation}
\begin{algorithm} [t!]
    \SetAlgoLined
     $\textbf{p}_{0}\gets \textbf{x}_{c0}$, $\textbf{p}_{i}\gets \textbf{x}_{q}(0)$, $\textbf{p}_{f}\gets \textbf{x}_{q}(T)$, $t_{f} \gets T$, $\boldsymbol{\tau}  \gets \{0\}$, $\boldsymbol{\mathcal{S}}^{c}\gets \emptyset$\\
     \While{true}{
        \If{generateCorridor($\textbf{p}_{0}$, $\textbf{p}_{f}$,$\mathcal{O}_{s}$)}{
            \If{not contain($\textbf{p}_{i}$,$\textbf{p}_{f}$, $\mathcal{S}$)}{\textbf{goto} Line 16}
            \uIf{$\textbf{p}_{f}=\textbf{x}_{q}(T)$}{ $\boldsymbol{\tau}$.pushback($t_{f}$), $\boldsymbol{\mathcal{S}}^{c}$.pushback($\mathcal{S}$);\\\textbf{return}\; }
            \Else{$\boldsymbol{\tau}$.pushback($t_{f}$), $\boldsymbol{\mathcal{S}}^{c}$.pushback($\mathcal{S}$),\\ $\textbf{p}_{0}\gets \textbf{p}_{f}, \textbf{p}_{i}\gets \textbf{p}_{f}, \textbf{p}_{f}\gets \textbf{x}_{q}(T)$;}
            \textbf{continue;}
        }
        $\textbf{p}_{f} \gets \textbf{x}_{q}(\frac{1}{2}(T+\boldsymbol{\tau}.\text{end}))$, $t_{f}\gets \frac{1}{2}(T+\boldsymbol{\tau}.\text{end})$
     }
     \textbf{return:}  $\boldsymbol{\tau}, \boldsymbol{\mathcal{S}}^{c}$
\caption{Visible and Safe Corridor Generation}
\label{alg:vsc_generation}
\end{algorithm}
We set $d_{\min}\geq \max({r_{qx},r_{qy},r_{qz}})+r_{c}$ to avoid collision with targets. 
\subsubsection{Safe Corridor Generation}
\label{subsubsec:chasing_sfc_generation}
We generate a safe flight corridor not only to avoid collisions with static obstacles $\mathcal{O}_{s}$ but also to ensure the visibility of the target. Algorithm \ref{alg:vsc_generation} describes the proposed safe flight corridor generation method. In \textbf{Line 3}, `generateCorridor' checks whether a polytope can be generated within the free space $\mathcal{F}_{s}$ enclosing the line segment connecting $\textbf{p}_{0}$ and $\textbf{p}_{f}$ \cite{sfc}. In \textbf{Line 4}, a function `contain' checks whether a part of $\textbf{x}_{q}(t)$, starting with $\textbf{p}_{i}$ and ending with $\textbf{p}_{f}$, lies within the corridor $\mathcal{S}$, generated in \textbf{Line 3}. 
Through Algorithm \ref{alg:vsc_generation}, we acquire a time-segment array $\boldsymbol{\tau}$ and an array of safe flight corridors $\boldsymbol{\mathcal{S}}^{c}$. Thanks to Casteljau algorithm \cite{bebot}, since the $\textbf{x}_{q}(t)$ and the primitives $\textbf{x}_{c}(t)$ are represented in Bernstein polynomial form, the divided trajectories are also expressed with Bernstein bases. According to $\boldsymbol{\tau}$, the trajectories are divided into $M$ polynomial segments with time intervals $[T_{0},T_{1}],\ldots,[T_{M-1},T_{M}]$, and the coefficients of split $\textbf{x}_{c}(t)$ and $\textbf{x}_{q}(t)$ are denoted as $[\textbf{C}_{1},\ldots,\textbf{C}_{M}]$ and $[\textbf{Q}_{1},\ldots,\textbf{Q}_{M}]$, respectively. 
\begin{proposition}    The maximum number of split trajectory segments ($M$) is two.
\end{proposition}
\begin{proof}
If the current position of the drone and the entire path of the targets lie in the same safe corridor, the functions, `genCorridor' and `contain', output true, and  $\textbf{p}_{f}=\textbf{x}_{q}(T)$. Then the Algorithm  \ref{alg:vsc_generation} terminates, and $M$ becomes 1. When \textbf{Line 12} is run, $\exists t \in [0,T],  \textbf{p}_{0}, \textbf{p}_{i} \in \textbf{x}_{q}(t)$ and $\textbf{p}_{f}=\textbf{x}_{q}(T)$. Since the prediction result $\textbf{x}_{q}(t)$ from Section. \ref{sec:target_trajectory_prediction} lies in $\mathcal{F}_{s}$, the functions in \textbf{Lines} \textbf{3}, \textbf{4}  and \textbf{7} output true. Then the while loop in Algorithm \ref{alg:vsc_generation} terminates, and $M$ becomes 2.
\end{proof} 
\subsubsection{Collision Check}
\label{subsubsec:collision_check}
To make the drone fly in the free space, $\mathcal{F}_{q}(t)$, we check whether flying trajectories satisfy (\ref{subeq:collision_between_static}): being confined in corridors and (\ref{subeq:collision_between_dynamic})-(\ref{subeq:collision_between_target}): avoiding collision with dynamic obstacles and targets. 
\begin{subequations}
    \label{eq:collision_check}
\begin{align}
    \label{subeq:collision_between_static}
    & \mathcal{V}(\textbf{C}_{k})\oplus \mathcal{E}(0,r_{c}^{-2}I_{3})\subset \mathcal{S}_{k}^{c},\ k=1,\ldots,M, \\
    \label{subeq:collision_between_dynamic}
    & \|\textbf{x}_{c}(t)-\textbf{x}_{o}(t)\|_{Q_{c}^{o}}^{2}- 1\geq0,\\
    \label{subeq:collision_between_target}
    & \|\textbf{x}_{c}(t)-\textbf{x}_{q}(t)\|_{Q_{c}^{q}}^{2}- 1\geq0
\end{align}
\end{subequations}
where $\mathcal{S}_{k}^{c}$ is the $k$-th corridor of $\boldsymbol{\mathcal{S}}^{c}$, and $Q_{o}^{c}$$=$$diag((r_{ox}+r_{c})^{-2},(r_{oy}+r_{c})^{-2},(r_{oz}+r_{c})^{-2})$ and $Q_{q}^{c}$$=$$diag((r_{qx}+r_{c})^{-2},(r_{qy}+r_{c})^{-2},(r_{qz}+r_{c})^{-2})$ are shape matrices of collision models that take into account the size of the drone, obstacles, and the targets. 
Since (\ref{eq:distance_target_divided}) is the sufficient condition for (\ref{subeq:collision_between_target}), for primitives that pass (\ref{eq:distance_target_divided}), we omit checking (\ref{subeq:collision_between_target}).
The primitives that pass tests in (\ref{eq:collision_check}) satisfy (\ref{subeq:drone_collision}).
\subsubsection{Visibility Check}
\label{subsubsec:visibility_check}
To prevent the target from being obscured by obstacles, we establish the following constraints. 
\begin{subequations}
    \label{eq:occlusion_check}
    \begin{align}
        \label{subeq:vis_static}
        & \mathcal{V}(\textbf{C}_{k}) \subset \mathcal{S}_{k}^{c},\ \forall 
 k=1,\ldots,M,\\
        \label{subeq:vis_dynamic}
        & \|\epsilon\textbf{x}_{c}(t)+(1-\epsilon)\textbf{x}_{q}(t)-\textbf{x}_{o}(t)\|_{Q_{o}}^{2}-1\geq 0,\forall \epsilon \in [0,1]
\end{align}
\end{subequations}
where $Q_{o}=diag(r_{ox}^{-2},r_{oy}^{-2},r_{oz}^{-2})$.
\begin{proposition} If (\ref{subeq:vis_static}) is satisfied, the targets are not occluded by static obstacles.
\end{proposition}
\begin{proof}
$\mathcal{S}_{k}^{c}$ contains the control points $\mathcal{V}(\textbf{Q}_{k})$ and is a convex polytope. Therefore, if (\ref{subeq:vis_static}) is satisfied, a convex hull composed of the control points $\mathcal{V}(\textbf{C}_{k})$ and $\mathcal{V}(\textbf{Q}_{k})$ belongs to $\mathcal{S}_{k}^{c}$, \textit{i.e.}, $\text{Conv}(\mathcal{V}(\textbf{C}_{k}), \mathcal{V}(\textbf{Q}_{k}))\subset \mathcal{S}_{k}^{c}$. Since $\mathcal{L}(\textbf{x}_{c}(t),\textbf{x}_{q}(t))$$\subset$ $\text{Conv}(\mathcal{V}(\textbf{C}_{k}),\mathcal{V}(\textbf{Q}_{k}))$ for $\forall t\in [T_{k-1}, T_{k}]$ and $\mathcal{S}_{k}^{c} \cap \mathcal{O}_{s} = \emptyset$, satisfaction of (\ref{subeq:vis_static}) makes $\mathcal{L}(\textbf{x}_{c}(t),\textbf{x}_{q}(t))\subset \mathcal{F}_{s}$, which means that targets are not occluded by static obstacles.
\end{proof}
(\ref{subeq:vis_dynamic}) represents a condition that \textit{Line-of-Sight} does not collide with dynamic obstacles, and the inequality in (\ref{subeq:vis_dynamic}) can be reformulated as (\ref{subeq:vis_dynamic_total}).
\begin{subequations}
    \label{eq:occlusion_check_dynamic}
\begin{align}
    \label{subeq:vis_dynamic_total}
    & \epsilon^{2}\sigma_{1}(t)+2\epsilon(1-\epsilon)\sigma_{2}(t)+(1-\epsilon)^{2}\sigma_{3}(t)\geq 0, \\
    \label{subeq:vis_dynamic_sub1}
    & \sigma_{1}(t) = \|\textbf{x}_{c}(t)-\textbf{x}_{o}(t)\|_{Q_{o}}^{2}- 1,\\
    \label{subeq:vis_dynamic_sub2}
    & \sigma_{2}(t) = (\textbf{x}_{c}(t)-\textbf{x}_{o}(t))^{\top}Q_{o}(\textbf{x}_{q}(t)-\textbf{x}_{o}(t))-1,\\
    \label{subeq:vis_dynamic_sub3}
    & \sigma_{3}(t) = \|\textbf{x}_{q}(t)-\textbf{x}_{o}(t)\|_{Q_{o}}^{2}-1
\end{align}
\end{subequations}
\begin{proposition}
    If $\sigma_{1}(t)$, $\sigma_{2}(t)$, $\sigma_{3}(t)$ $\geq 0$, $\forall t \in [0,T]$, the targets are not occluded by dynamic obstacles.
\end{proposition}
\begin{proof}
    $\epsilon^{2}$, $2\epsilon(1-\epsilon)$, and $(1-\epsilon)^{2}$ are nonnegative for $\forall \epsilon \in [0,1]$. Therefore, if $\sigma_{1}(t)$, $\sigma_{2}(t)$ and $\sigma_{3}(t)$ are nonnegative for $\forall t\in[0,T]$, then $[\epsilon^{2},2\epsilon(1-\epsilon),(1-\epsilon)^{2}][\sigma_{1}(t),\sigma_{2}(t),
    \sigma_{3}(t)]^{\top}$ becomes nonnegative  for $\forall t\in[0,T]$, and 
    \text{($\ref{subeq:vis_dynamic_total}$)} gets satisfied.
\end{proof}
Nonnegativeness of (\ref{subeq:vis_dynamic_sub1}) and (\ref{subeq:vis_dynamic_sub3}) are necessary conditions for  (\ref{eq:distance_target_divided}) and (\ref{subeq:collision_between_dynamic_target}), respectively.
The prediction results $\textbf{x}_{q}(t)$ naturally satisfies $\sigma_{3}\geq0$, so for efficient checks, it is enough to check Bernstein coefficients in (\ref{subeq:vis_dynamic_sub2}) for primitives that pass the test (\ref{eq:distance_target_divided}).

Furthermore, to observe the $i$-th and $j$-th targets simultaneously with limited camera FOV $\theta_{f}$, we formulate a constraint as follows:
\begin{equation}
    \label{eq:fov_check}
    \begin{aligned}
        &\|\textbf{x}_{c}(t)-\textbf{x}_{f}^{i,j}(t) \|_{diag(1,1,0)}^{2}-(r_{f}^{i,j}(t))^{2}\geq 0,\\
        &r_{f}^{i,j}(t) = \frac{1}{2\sin{\theta_{f}}}\|\textbf{x}_{q}^{i}(t)-\textbf{x}_{q}^{j}(t)\|_{2},\\
        &\textbf{x}_{f}^{i,j}(t)= F_{q}^{i,j}(\textbf{x}_{q}^{i}(t)-\textbf{x}_{q}^{j}(t))+\textbf{x}_{q}^{j}(t),\\
        &F_{q}^{i,j}= \frac{1}{2}\begin{bmatrix}
            1 & (-1)^{k_{f}^{i,j}}\cot{\theta_{f}} & 0 \\
            (-1)^{k_{f}^{i,j}+1}\cot{\theta_{f}} & 1 & 0 \\
            0 &0 &1
        \end{bmatrix}
    \end{aligned}
\end{equation}
where $k_{f}^{i,j}$ is a function of $\textbf{x}_{q}^{i}(0)$ and $\textbf{x}_{q}^{j}(0)$. A cylinder with center at $\textbf{x}_{f}^{i,j}(t)$ and radius of $r_{f}^{i,j}(t)$ includes the \textit{Deadzone} where the drone cannot simultaneously see the two targets. Please refer to \cite{qp_chaser} for the details.
\begin{figure}[t!]
    \centering
    \begin{subfigure}[t]{0.24\textwidth}
    \centering \includegraphics[width=\textwidth]{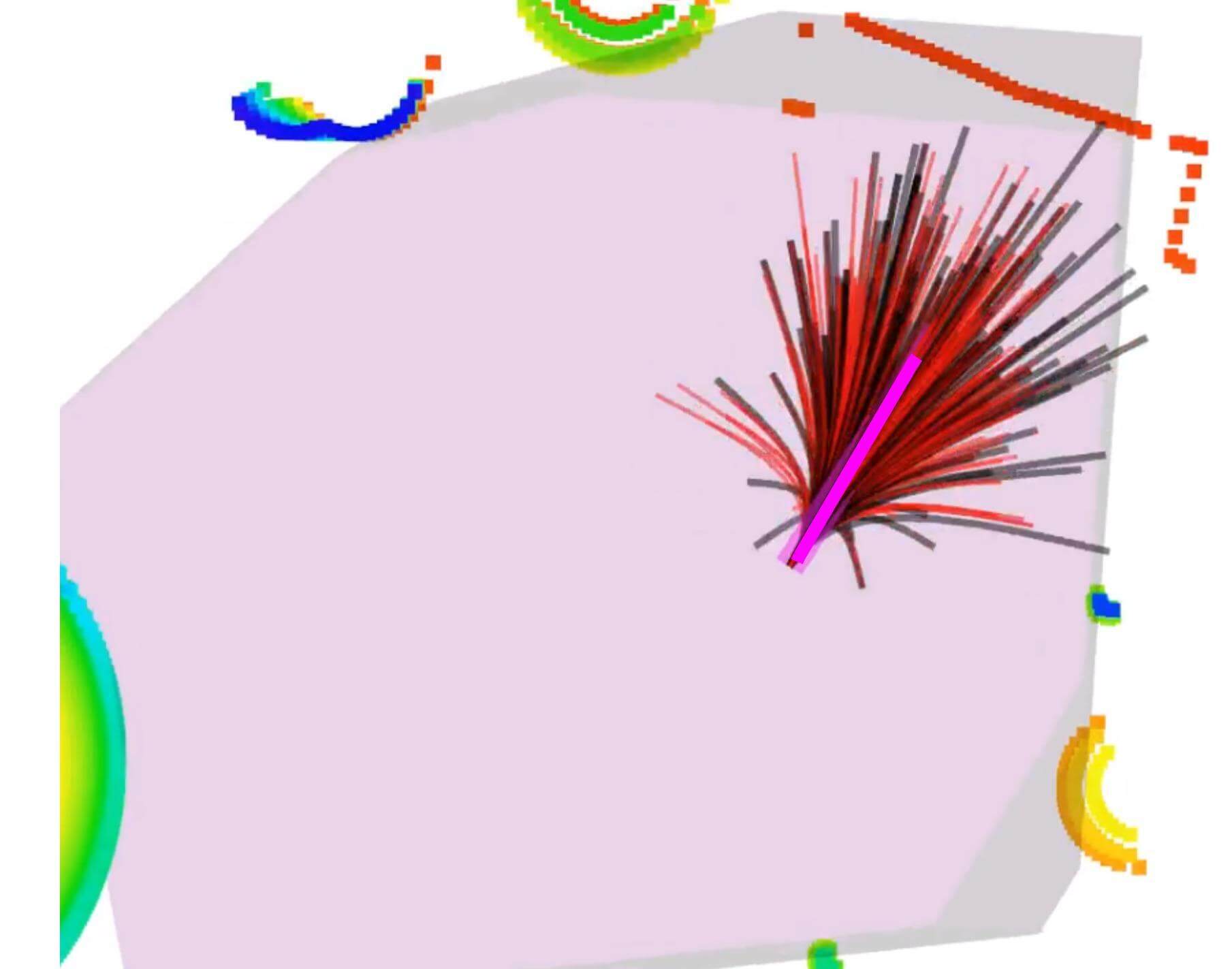}
    \caption{Target prediction}
    \label{subfig:target_prediction}
    \end{subfigure}
    \begin{subfigure}[t]{0.24\textwidth}
    \centering \includegraphics[width=\textwidth]{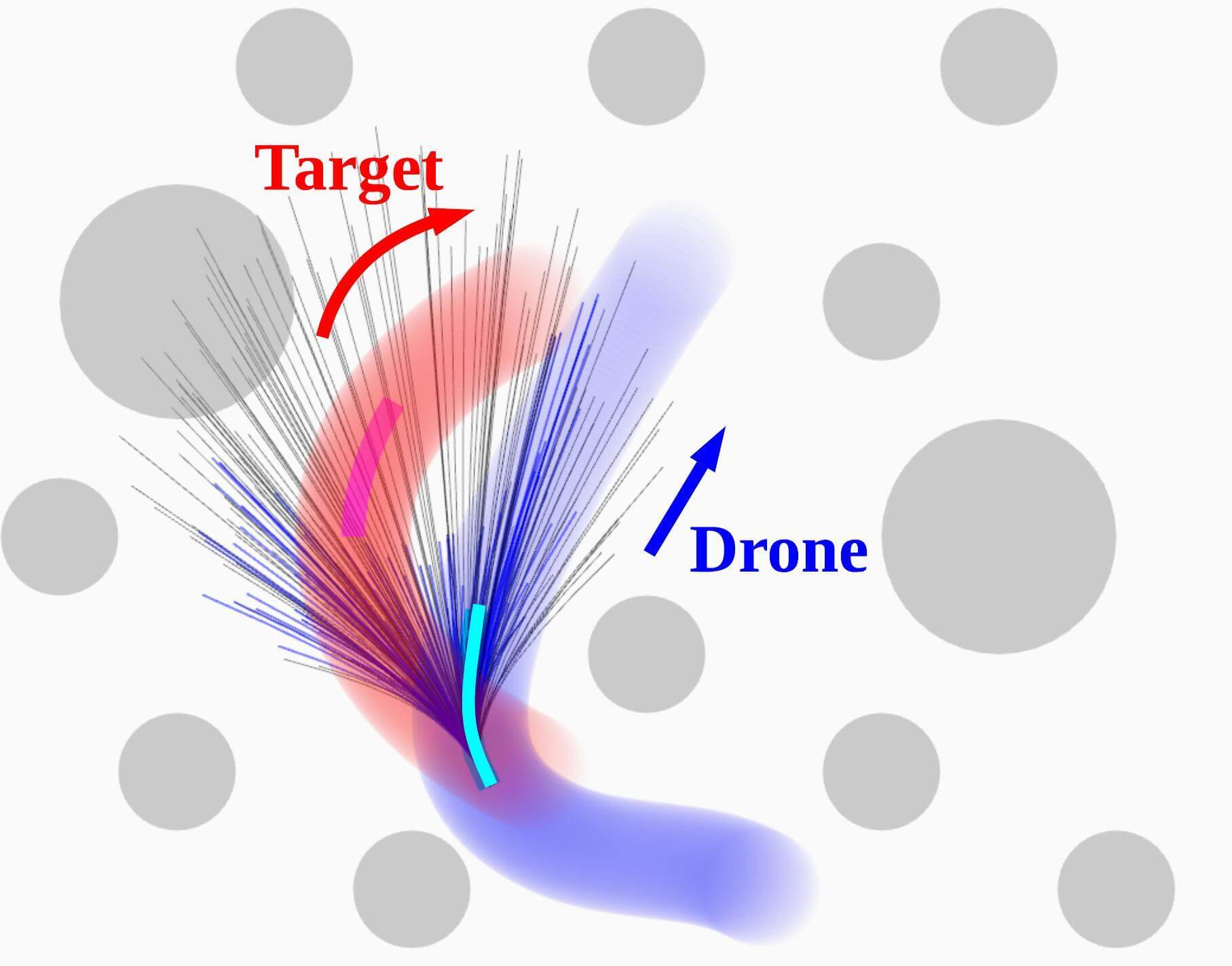}
    \caption{Trajectory planning}
    \label{subfig:drone_planning}
    \end{subfigure}
    \caption{\{Black, red, magenta\} and \{black, blue, cyan\} splines represent \{primitives, feasible and the best trajectories\} in the prediction and planning, respectively. The pink-shaded region on the left is a generated safe corridor ($\mathcal{S}^{q}$) in point cloud, and grey cylinders on the right are obstacles.} 
    \label{fig:result_example}
\end{figure}


\subsubsection{Dynamical Limit Check}
\label{subsubsec:dynamical_limit_check}
To ensure the drone does not exceed dynamic limits (\ref{eq:dynamical_limits}), we verify whether its primitives satisfy the following inequalities.
\begin{subequations}
    \label{eq:dynamical_limit_translation}
    \begin{align}
        &v_{\max}^{2}-\|\dot{\textbf{x}}_{c}(t)\|_{2}^{2} \geq 0,\\
        &a_{\max}^{2}-\|\ddot{\textbf{x}}_{c}(t)\|_{2}^{2} \geq 0
    \end{align}
\end{subequations}
Additionally, when the drone directly observes a single target or stares at points amidst multiple targets, the following constraints can be formulated to constrain the yaw rate.
\begin{subequations}
    \label{eq:dynamical_limit_rotation}
    \begin{align}
    \label{subeq:dynamical_limit_rotation1}
    &\dot{\psi}(t) =
    \frac{(\textbf{x}_{q}(t)-\textbf{x}_{c}(t))\times(\dot{\textbf{x}}_{q}(t)-\dot{\textbf{x}}_{c}(t))\cdot{[0 \ 0 \ 1]^{\top}}}{\|\textbf{x}_{q}(t)-\textbf{x}_{c}(t)\|_{diag(1,1,0)}^{2}},\\
    \label{subeq:dynamical_limit_rotation2}
    & -\dot{\psi}_{\max} \leq \dot{\psi}(t) \leq \dot{\psi}_{\max}
    \end{align}
\end{subequations}
Since the numerator and denominator of the right-hand side of (\ref{subeq:dynamical_limit_rotation1}) are both Bernstein polynomials, the right-hand side takes the form of a rational Bernstein polynomial \cite{bebot}, the ratio of coefficients between the numerator and denominator is calculated and compared with the bound $[-\dot{\psi}_{max},\dot{\psi}_{max}]$.
\subsection{Best Chasing Trajectory Selection}
\label{subsec:best_trajectory_selection}
Among the primitives that pass (\ref{eq:distance_target_divided})-(\ref{eq:dynamical_limit_rotation}), we select the best chasing trajectory. We evaluate the following metric, which consists of a cost penalizing high-order derivatives of trajectory and a cost to maintain appropriate relative distance from the targets.
\begin{subequations}
    \label{eq:chasing_cost}
    \begin{align}
        \min \ &J_{1}+J_{2}, \\
        &J_{1} = \int_{0}^{T}w_{a}\|\ddot{\textbf{x}}_{c}(t)\|_{2}^{2}+w_{j}\|\dddot{\textbf{x}}_{c}(t)\|_{2}^{2}dt,\\
        &J_{2} = \sum_{i=1}^{N_{q}} \int_{0}^{T}(\|\textbf{x}_c(t)-\textbf{x}_{q}^{i}(t)\|_{2}^{2}-d_{des}^{2})^{2}dt
    \end{align}
\end{subequations}
where $w_{a}$ and $w_{j}$ are weight factors, and $d_{des} = \frac{1}{2} (\bar{r}_{c}+\underbar{$r$}_{c})$. The primitive having the minimum metric is selected as a final trajectory. Since all integrands in (\ref{eq:chasing_cost}) are Bernstein polynomials, the metric can be calculated with a few arithmetic operations with coefficients \cite{bebot}. This allows us to select the best trajectory quickly. An example of trajectory planning is shown in Fig. \ref{subfig:drone_planning}.

\section{Experiments}
\label{sec:experiments}
In this section, the presented method is validated through various target-following settings. We measure the distance between the drone and environments, defined as (\ref{subeq:minimum_distance}), to assess drone safety and the metric defined as (\ref{subeq:visibility_score}) to evaluate target visibility. 
\begin{subequations}
    \label{eq:performance_metric}
    \begin{align}
        \label{subeq:minimum_distance}
        &\chi(\textbf{x}_{c}(t);\mathcal{O}(t),\mathcal{T}(t)) = \min_{\substack{\textbf{x}\in\mathcal{E}(\textbf{x}_c(t), r_{c}^{-2}I)\\\textbf{y}\in \mathcal{O}(t)\cup \mathcal{T}(t)}} \|\textbf{x}-\textbf{y}\|_{2}\\
        \label{subeq:visibility_score}
        &\phi(\textbf{x}_{c}(t);\textbf{x}_{q}^{1:N_{q}}(t),\mathcal{O}(t))= \min_{i \in\{1:N_{q}\}} \min_{\substack{\textbf{x}\in \mathcal{L}(\textbf{x}_{c}(t),\textbf{x}_{q}^{i}(t))\\ \textbf{y}\in \mathcal{O}(t)}}\|\textbf{x}-\textbf{y}\|_{2}
    \end{align}
\end{subequations}
With the above performance indices, we validate the operability of the proposed planner for four scenarios: 1) single-, 2) dual-target following in unstructured environments, target following amidst 3) circular (2D), and 4) spherical (3D) dynamic obstacles, through simulation and hardware demonstration. 
The details of tests are explained in the following subsections, and Table \ref{tab:validations} summarizes the reported performance, indicating no occlusion and collision during the tests. In addition, we compare the performance with optimization-based planners and investigate the performance in multi-target scenarios.

Both positions and velocities of moving objects are provided in the simulators in Sections \ref{subsec:simulations_dense_dynamic_obstacles}, \ref{subsec:benchmark_test}, and \ref{subsec:multi_target_tracking_tests}, while we acquire the positional information using camera sensors or motion capture systems and estimate the velocities of moving objects using extended Kalman filters in other validations.
The number of samples in both prediction and planning is set to 1000, and four threads are utilized in parallel computation in all validations. We use a computer with an Intel i7 10th-gen CPU and 16GB RAM and an edge device with an NVIDIA Carmel ARM®v8.2 CPU and 8GB RAM.
%
\subsection{Simulations in Large-scale Environments}
\label{subsec:simulations_large_scale}
\begin{figure}[t!]
\centering
\includegraphics[width=\linewidth]{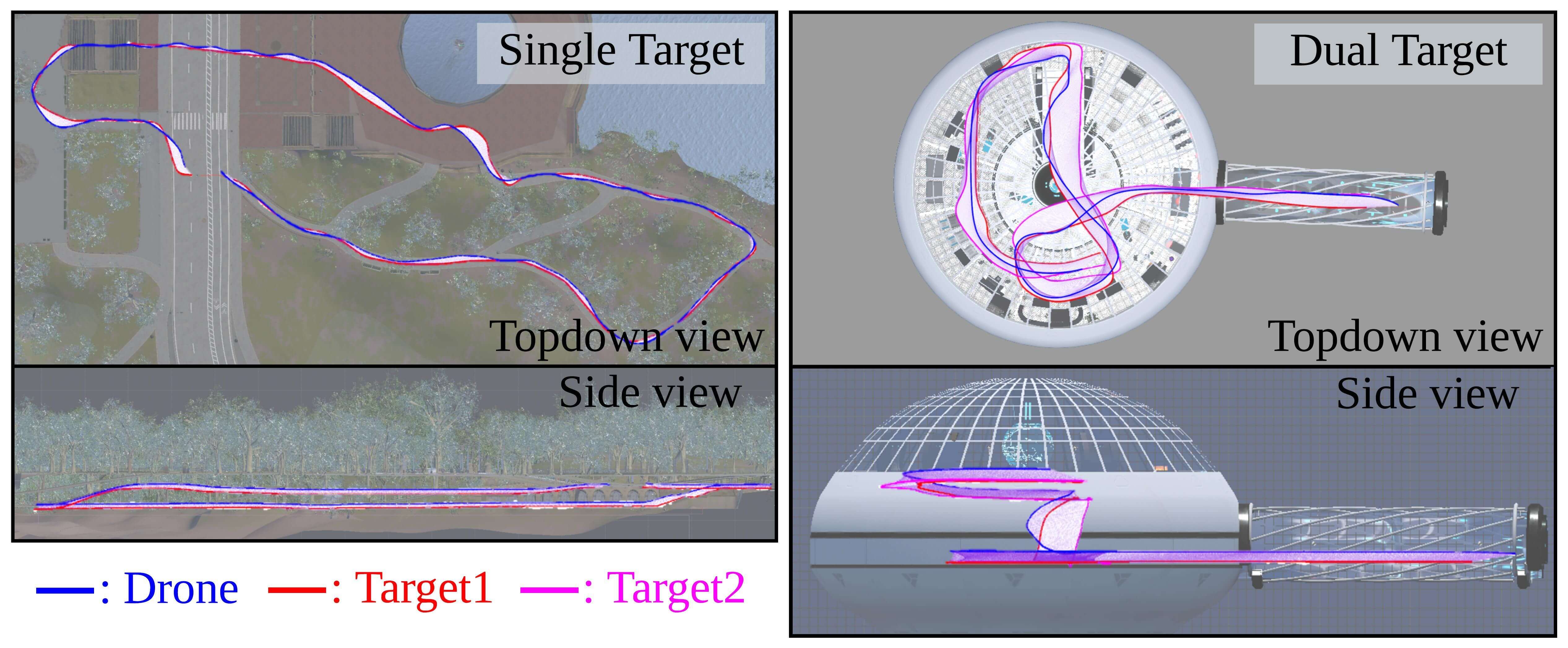}
\caption{Simulation results in a city park (left, single target) and a futuristic building (right, dual target). The position histories of the drone (blue) and targets (red, magenta) are plotted.}
\label{fig:mp_chaser_airsim_single_dual}
\end{figure}

We perform missions to follow single and dual targets among unstructured environments in AirSim \cite{airsim} simulator.
A man walks 392 m in a city park, and two running robots ascend from the lower level to the upper level of a dome-shaped building. The planning results are shown in Fig. \ref{fig:mp_chaser_airsim_single_dual}. Despite the high computational demands imposed by the physics engine, the whole pipeline is completed within 30 milliseconds.
\begin{table}[t]
\centering
\caption{Reported metrics in validations  (min/mean/max)}
\begin{tabular}{@{}lcc@{}}
\toprule \midrule
Scenarios      &Safety metric [m] &Visibility metric [m]        \\ \midrule[0.4pt]
\multicolumn{1}{l}{Unstructured-Single-Sim} & 0.47/2.20/4.42 & 0.02/2.32/7.84 \\
\multicolumn{1}{l}{Unstructured-Dual-Sim}  & 0.59/2.14/4.40 & 0.04/1.28/3.25 \\
\multicolumn{1}{l}{Dynamic-2D-Sim}  & 0.05/0.13/0.25 & 0.13/0.27/0.48 \\
\multicolumn{1}{l}{Dynamic-3D-Sim}  & 0.05/0.16/0.70 & 0.29/0.37/0.52 \\
\multicolumn{1}{l}{Unstructured-Single-Exp} & 0.19/1.09/2.95 & 0.31/0.79/1.78 \\
\multicolumn{1}{l}{Unstructured-Dual-Exp}  & 0.59/2.14/4.40 & 0.03/0.65/0.88 \\
\multicolumn{1}{l}{Dynamic-2D-Exp}  & 0.05/0.13/0.25 & 0.22/0.94/2.10 \\
\multicolumn{1}{l}{Dynamic-3D-Exp}  & 0.13/0.46/0.97 & 0.27/1.10/1.98 \\
\bottomrule
\end{tabular}
\label{tab:validations}
\end{table}
\begin{figure}[t!]
\centering
\begin{subfigure}[t!]{0.42\textwidth}
\centering \includegraphics[width=\textwidth]{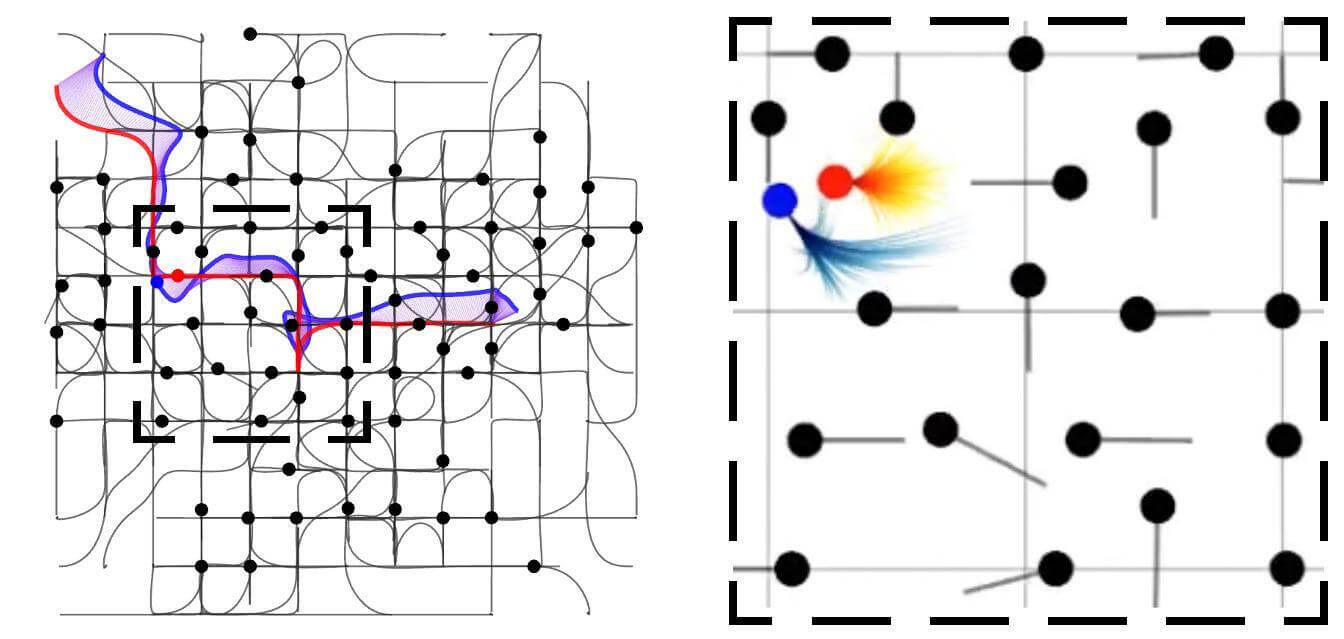}
\caption{2D Environments}
\label{subfig:2d_simulation}
\end{subfigure}
\begin{subfigure}[b!]{0.42\textwidth}
\centering \includegraphics[width=\textwidth]{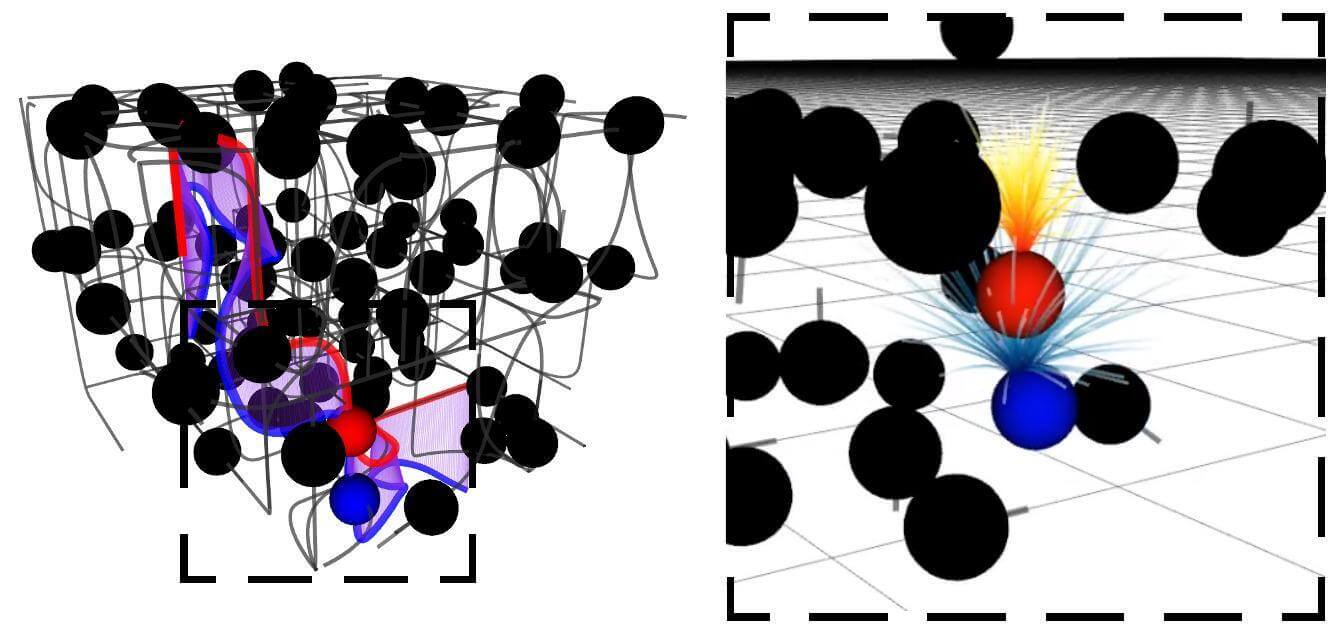}
\caption{3D Environments}
\label{subfig:3d_simulation}
\end{subfigure}
\caption{Chasing results amidst 69 dynamic-obstacles (black). Blue, red and grey splines represent the position histories of the drone, the target and the obstacles, respectively.}
\label{fig:dynamic_obstacle_simulation}
\vspace{-4mm}
\end{figure}
\subsection{Simulations in Dense Dynamic Obstacle Spaces}
\label{subsec:simulations_dense_dynamic_obstacles}
We test the proposed planner in dense dynamic-obstacle environments. Target moves in a $6\times6 \ \text{m}^{2}$ 2D space and a $3 \times 3\times 2 \ \text{m}^{3}$ 3D space. In both environments, 69 dynamic obstacles with radius 0.07 m wander around with the maximum speed 0.93 m/s and 0.92 m/s. In 2D environments, $r_{oz}, r_{qz}=\infty$, so feasible movements of the drone must be found only on the $x$-$y$ plane, which makes a mission more challenging. Fig. \ref{fig:dynamic_obstacle_simulation} shows simulation results. Despite numerous obstacles, the prediction and planning are done in 50 Hz.
\subsection{Flight Test in Cluttered Environments}
\label{subsec:flight_test_cluttered}
\begin{figure}[t!]
\centering
\includegraphics[width=\linewidth]{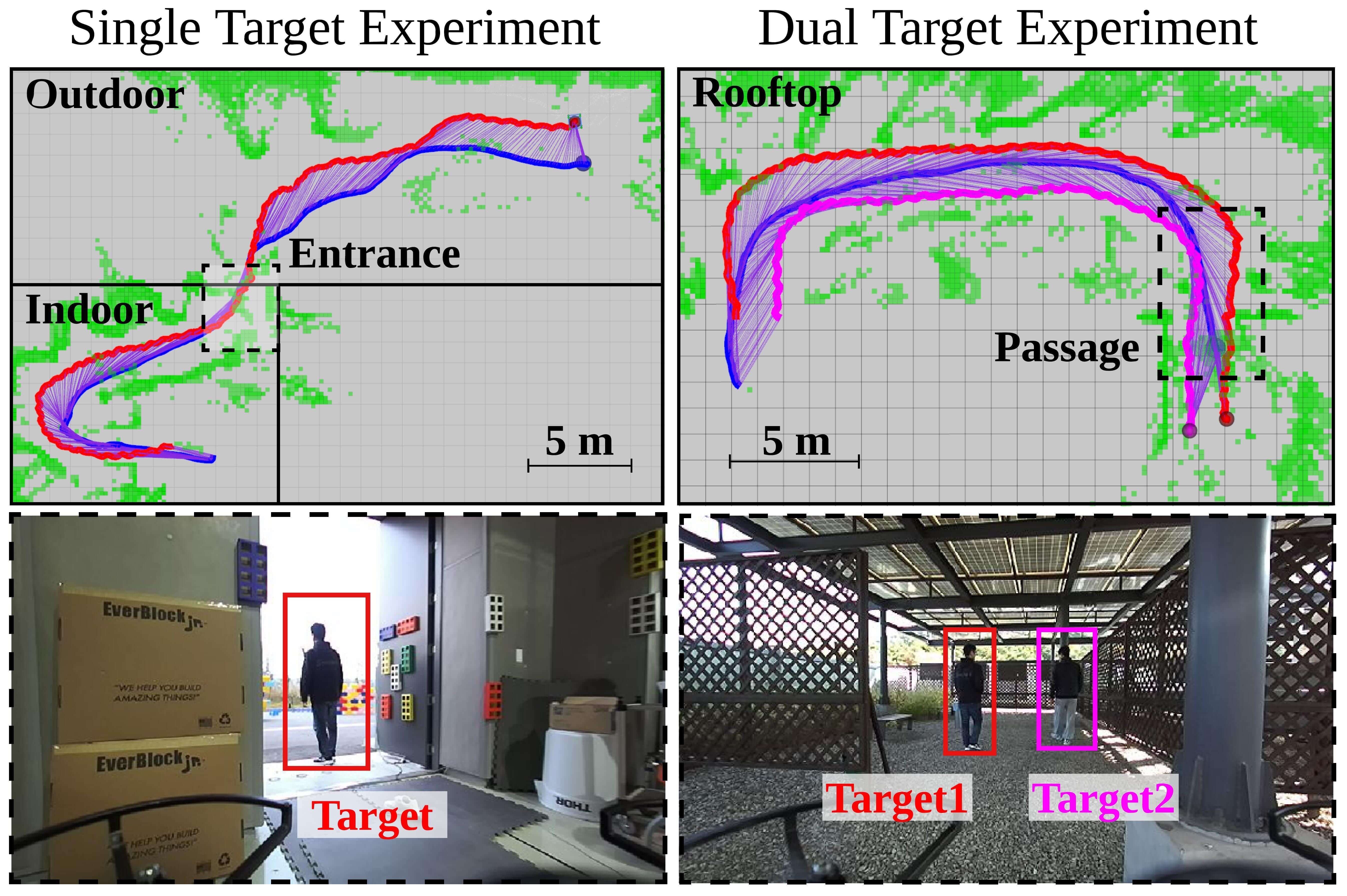}
\caption{Human target experiments. The reported position histories of the drone (blue) and targets (red, magenta) are plotted in a top-down view. Green-colored regions and a bundle of purple segments are occupied space and \textit{Line-of-Sights} between the targets and the drone, respectively.}
\label{fig:mp_chaser_exp}
\end{figure}

We perform real-world experiments with human targets. Jetson Xavier NX is equipped as an onboard computer, and a ZED2 \cite{zed} stereo camera and Pixhawk4, a flight controller, are mounted on the drone. In the first scenario, one person starts indoors and moves outside, while in the second, two men walk among structures on a building rooftop. Fig. \ref{fig:mp_chaser_exp} summarizes the experiment results. Despite the simultaneous execution of other submodules such as localization, mapping, and object detection, the proposed method completes its calculations within 30 milliseconds.
\begin{figure}[t!]
\centering
\begin{subfigure}[t]{0.23\textwidth}
\label{figure:crazyflie3D}
\centering \includegraphics[width=1.0\linewidth]{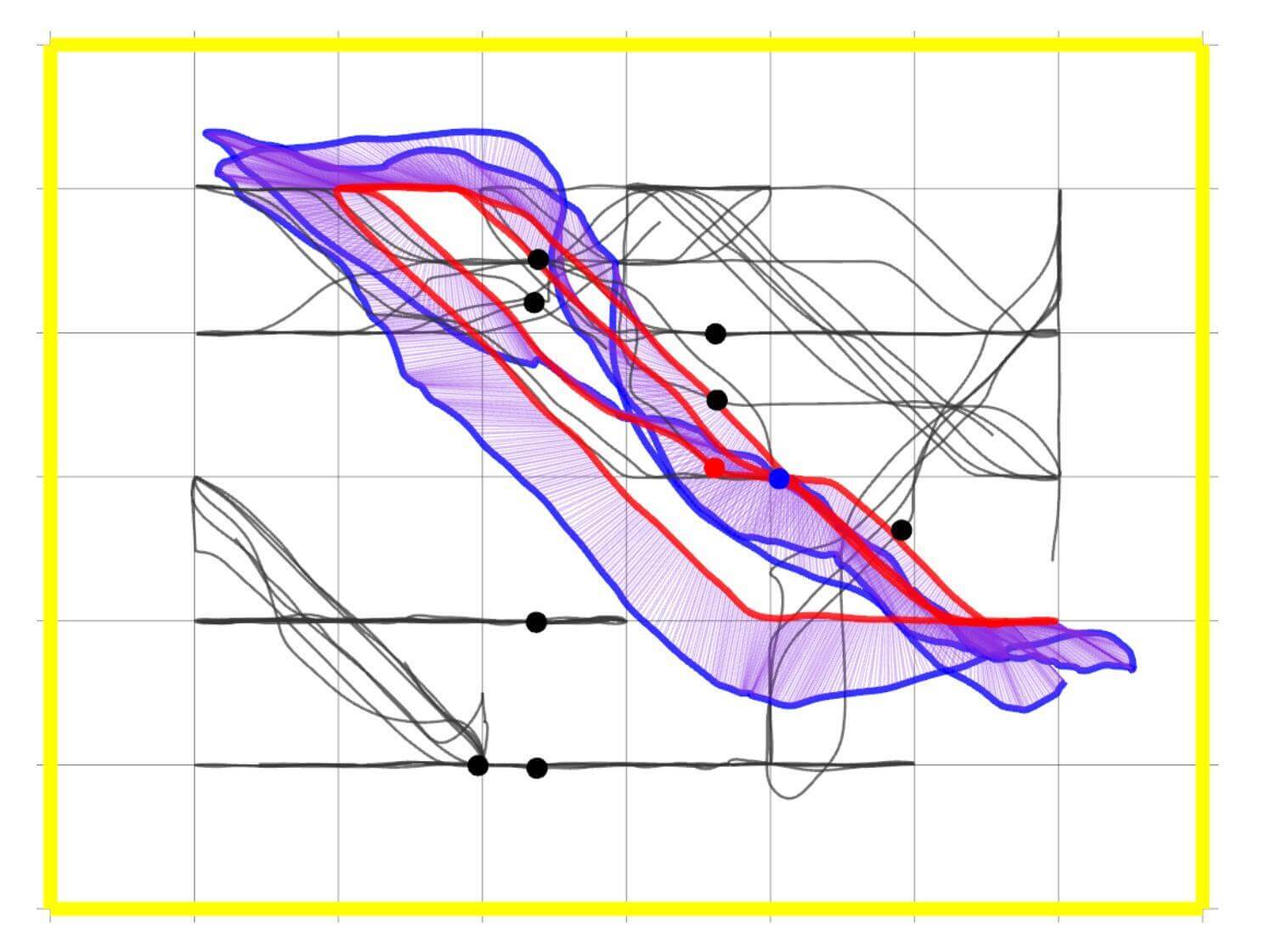}
\caption{Flights in a 2D space}
\end{subfigure}
\begin{subfigure}[t]{0.24\textwidth}
\label{figure:crazyflie2D}
\centering \includegraphics[width=1.0\linewidth]{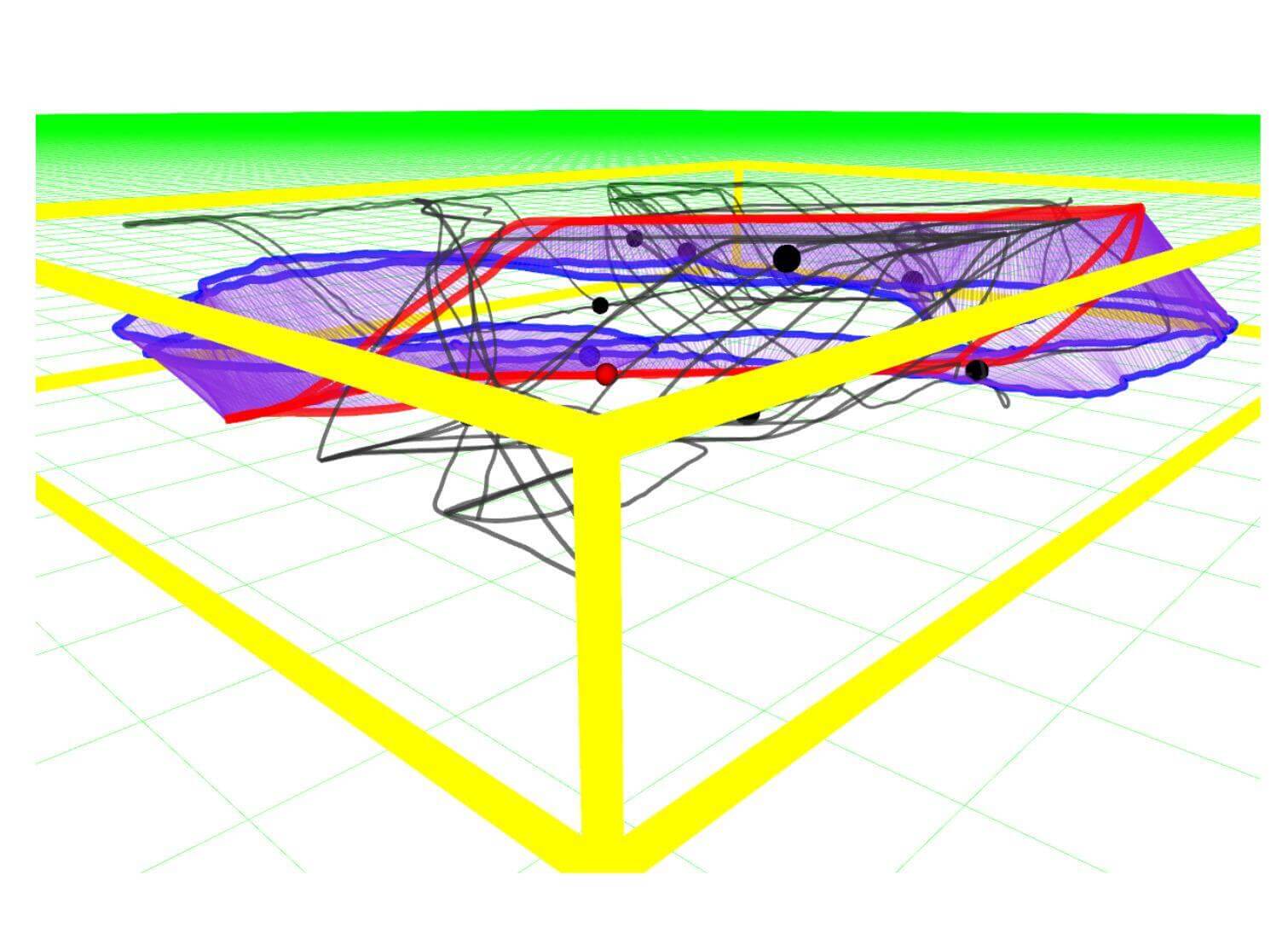}
\caption{Flights in a 3D space}
\end{subfigure}
\caption{Total flight histories of a single chaser (blue), a single target (red) and eight dynamic obstacles (black). The yellow lines represent the confined flight space.}
\label{fig:craziflie}
\end{figure}
\subsection{Flight Test in Dynamic Obstacle Spaces}
\label{subsec:flight_test_dynamic_obstacles}
For the hardware demonstration of the dynamic obstacle scenarios, we used ten Crazyflie 2.1 quadrotors. One serves as the chaser drone, another as the target, while the rest play the roles of dynamic obstacles. All drones move in an $8\times6$ $\text{m}^{2}$ space in the first scenario (2D) and in an $8 \times 6 \times 1\  \text{m}^{3}$ space in the second scenario (3D) as shown in Fig. \ref{fig:thumbnail}. Optitrack motion capture system is employed to measure positions of the quadrotors, and Crazyswarm \cite{crazyflie} broadcasts control commands calculated on a laptop to the quadrotors. Fig. \ref{fig:craziflie} shows the reported results in flight tests. The planning results are updated every 100 milliseconds, but the computation of the entire pipeline takes less than 10 milliseconds.
\subsection{Benchmark Tests}
\label{subsec:benchmark_test}
\begin{figure}[t!]
\centering\includegraphics[width=0.8\linewidth]{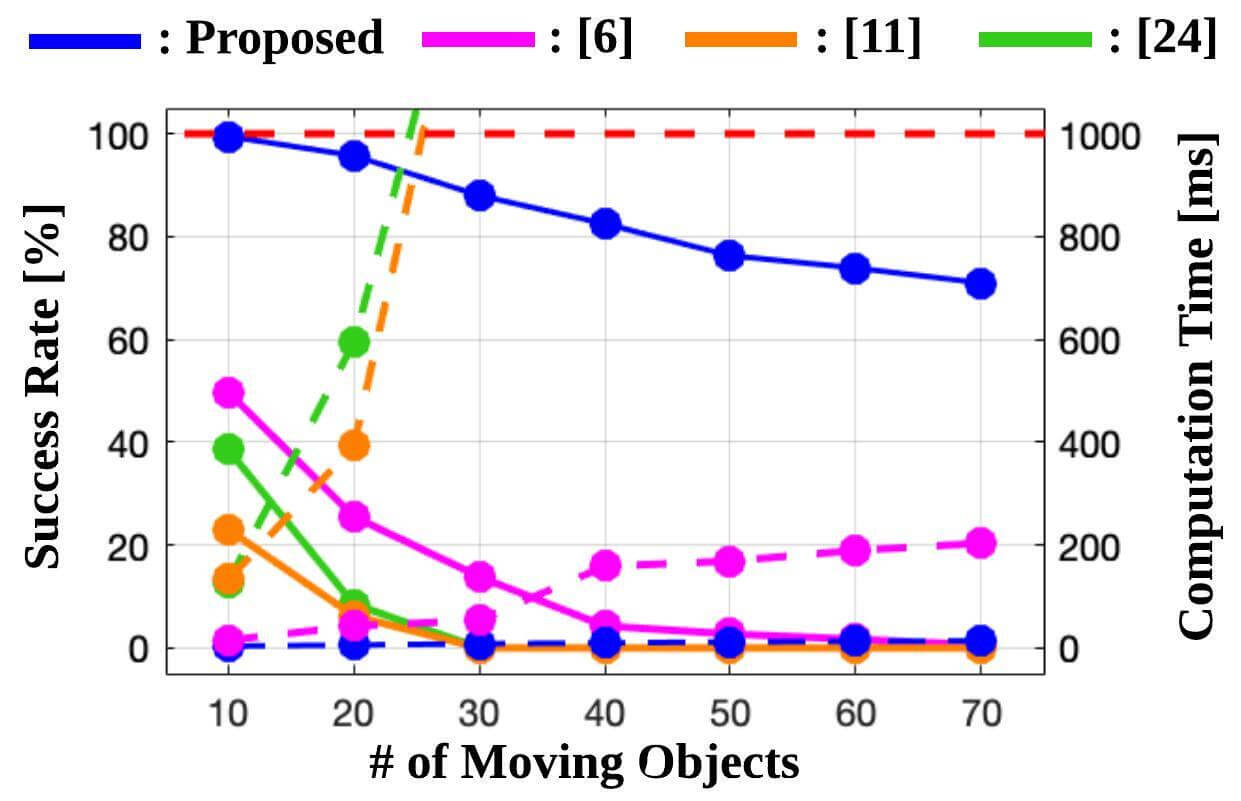}
\caption{Performance comparison  in $6\ \times 6\ \text{m}^{2}$ 2D dynamic environments. Solid lines and dashed lines represent success rate and computation time, respectively. The red dashed line means planning horizon ($1\ \text{second}$). }
\vspace{-4mm}
\label{fig:benchmark_comparison}
\end{figure}

The performance of the proposed planner is investigated by comparing it with three optimization-based chasing planning algorithms in dynamic environments: \cite{bonatti}, \cite{nageli}, and \cite{penin}. Ten to seventy cylindrical objects with a radius of 0.07 m move in the same space in Fig. \ref{subfig:2d_simulation}. One of them is the target, while the others are obstacles, and they move at the maximum speed of 1 m/s. We ran 1000 tests for each number of moving objects, varying their paths and the initial position of the tracker. We define success as no occlusion or collision occurring until all objects come to a stop at the end of the pre-defined time period. We then measure the success rate and average computation time for each test, and Fig. \ref{fig:benchmark_comparison} shows the results. As the number of moving obstacles in the confined space increases, so does the number of cases where obstacles 1) rush into the tracker and 2) cut in between the tracker and the target, resulting in challenging conditions to satisfy all requirements: (\ref{eq:distance_target})-(\ref{eq:yaw_rate}). Also, the calculation time to generate the tracking trajectory increases.
In our planner’s case, among sample-check-select processes, only the number of conditions to check increases linearly, without changes in the sample and select processes. This results in linear increases in computation time. Fast planning update helps our planner achieve higher success rates than other methods in dense and rapidly changing dynamic environments.
\subsection{Multi-Target Tracking Test}
\label{subsec:multi_target_tracking_tests}
\begin{figure}[t!]
\centering\includegraphics[width=0.75\linewidth]{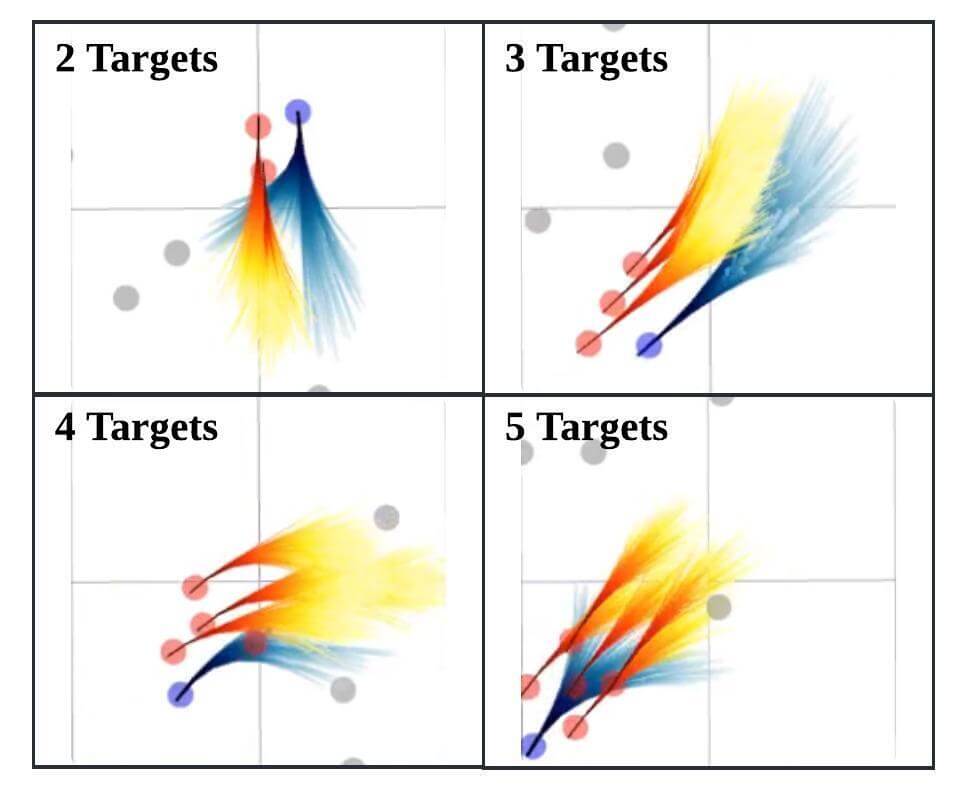}
\caption{Multi-target tracking in dynamic environments. The tracker (blue) follows the targets (red) amidst obstacles (grey).}
\label{fig:multi_target_sim}
\end{figure}

We conduct multi-target tracking missions in dynamic environments The number of dynamic obstacles is set to 9, and their moving paths are the same in the benchmark tests. Two, three, four, and five targets move among obstacles, and the relative distance among them varies from 0.2 m to 0.6 m, as shown in Fig. \ref{fig:multi_target_sim}. We set the number of primitives for each target to 1000, and we measure the success rate and the computation time and summarize in Table \ref{tab:multi_target}. As the number of targets increases, the computation time increases linearly; however, it is sufficiently fast for use in dynamic environments. On the other hand, the success rate decreases because the increase in the number of targets makes it difficult to satisfy the distance condition (\ref{eq:distance_target_divided}).
However, the main reason for failures is dynamic obstacles crossing among targets, which can cause inevitable occlusions that cannot be resolved by a single chasing drone.
\section{Conclusion}
\label{sec:conclusion}
We presented the trajectory planning pipeline for aerial tracking. The pipeline consists of the target motion prediction and chasing trajectory planning. Both algorithms are based on the sample-check-select strategy and leverage the properties of the Bernstein polynomial for efficient computation. The prediction module successfully forecasts the trajectories of the targets not to traverse static and dynamic obstacle spaces. The planning module generates trajectories considering target visibility (occlusion, camera FOV), collision and dynamic limits (both translational and rotational rates). Various validations, including scenarios involving dynamic-obstacle environments and multi-target following, confirm the efficiency and effectiveness of the proposed system. 
\begin{table}[t]
\centering
\caption{Multi-Target Tracking Performance}
\begin{tabular}{@{}lcccc@{}}
\toprule \midrule
The Number of Targets      &2   &3   &4   &5     \\ \midrule[0.4pt]
\multicolumn{1}{l}{Success Rate [\%]} & 89.0 & 82.5 & 74.5 & 72.9 \\
\multicolumn{1}{l}{Computation Time [ms]}  & 5.67 & 7.68 & 9.85 & 11.84\\
\bottomrule
\end{tabular}
\label{tab:multi_target}
\vspace{-4mm}
\end{table}

\end{document}